\documentclass[conference]{IEEEtran}
\IEEEoverridecommandlockouts
\usepackage{cite}
\usepackage{amsmath,amssymb,amsfonts}
\usepackage{algorithmic}
\usepackage{graphicx}
\usepackage{textcomp}
\usepackage{xcolor}
\def\BibTeX{{\rm B\kern-.05em{\sc i\kern-.025em b}\kern-.08em
    T\kern-.1667em\lower.7ex\hbox{E}\kern-.125emX}}
\usepackage{soul}
\usepackage{booktabs}
\usepackage{multirow}
\usepackage{tcolorbox}
\usepackage{enumitem}
\usepackage{subcaption}
\usepackage{hyperref}

\newcommand{\citeTodo}[1]{{\color{red}[??]}}

\newcommand{\COMMENT}[1]{}
\newboolean{showcomments}
\setboolean{showcomments}{true}         
\ifthenelse{\boolean{showcomments}}
  {\newcommand{\nb}[2]{
  \fbox{\bfseries\sffamily\scriptsize#1}
     {\sf\small$\blacktriangleright$\textit{\textcolor{red}{#2}}$\blacktriangleleft$}
   }
  }
  {\newcommand{\nb}[2]{}
   
  }

\newcommand\anybotics[1]{{#1}}

\begin{document}

\title{\anybotics{Bridging Research and Practice in Simulation-based Testing of Industrial Robot Navigation Systems}}

\author{
\IEEEauthorblockN{Sajad Khatiri\IEEEauthorrefmark{1},
\anybotics{Francisco Eli Viña Barrientos}\IEEEauthorrefmark{2}, 
\anybotics{Maximilian Wulf}\IEEEauthorrefmark{3}, 
Paolo Tonella\IEEEauthorrefmark{4}, 
Sebastiano Panichella\IEEEauthorrefmark{5}}
\IEEEauthorblockA{\IEEEauthorrefmark{1}\IEEEauthorrefmark{4}Università della Svizzera italiana, Lugano, Switzerland\\}
\IEEEauthorblockA{\IEEEauthorrefmark{1}\IEEEauthorrefmark{5}University of Bern, Bern, Switzerland \\}
\IEEEauthorblockA{\IEEEauthorrefmark{2}\IEEEauthorrefmark{3}\anybotics{ANYbotics AG, Zurich, Switzerland} 
}
sajad.mazraehkhatiri@unibe.ch, \anybotics{\{fvina, mwulf\}@anybotics.com}, paolo.tonella@usi.ch, sebastiano.panichella@unibe.ch
}

\maketitle

\begin{abstract}
Ensuring robust robotic navigation in dynamic environments is a key challenge, as traditional testing methods often struggle to cover the full spectrum of operational requirements. This paper presents the industrial adoption of Surrealist, a simulation-based test generation framework originally for UAVs, now applied to the ANYmal quadrupedal robot for industrial inspection. Our method uses a search-based algorithm to automatically generate challenging obstacle avoidance scenarios, uncovering failures often missed by manual testing.
In a pilot phase, generated test suites revealed critical weaknesses in one experimental algorithm (40.3\% success rate) and served as an effective benchmark to prove the superior robustness of another (71.2\% success rate). The framework was then integrated into the ANYbotics workflow for a six-month industrial evaluation, where it was used to test five proprietary algorithms. A formal survey confirmed its value, showing it enhances the development process, uncovers critical failures, provides objective benchmarks, and strengthens the overall verification pipeline.

\end{abstract}

\begin{IEEEkeywords}
Robotic Navigation, Local Planning, Obstacle Avoidance, Simulation Environments, Search-Based Testing, Environment Generation, Quadrupedal Robots 
\end{IEEEkeywords}

\section{Introduction}
\label{sec:introduction}
Robots are revolutionizing industrial workflows in manufacturing, logistics, inspection, and maintenance by improving efficiency, productivity, and safety~\cite{Ingrand19,app13042304}. Mobile autonomous robots are key for operating in complex, dynamic environments and handling hazardous or impractical tasks~\cite{HUANG20191,3564696,Torras24,KebedeGAH24,Gehring2021ANYmalRobot}. This trend demands robust, reliable platforms for safe real-world deployment~\cite{GUIOCHET201743,surrealist}.
Autonomous navigation \cite{3727642}—the ability to plan and follow paths—is vital for industrial robots. Obstacle avoidance enables safe interaction with static and dynamic hazards \cite{44033}, including equipments, vehicles, and humans \cite{8107677,8194898,VRstudy}. Failures can lead to collisions, damage, downtime, and safety risks \cite{DiSorboTOSEM2023,ZampettiKPP22}, underscoring the need for rigorous testing \cite{surrealist,aerialist,birchler2024roadmapsimulationbasedtestingautonomous,BirchlerRKGLHKP23}. 

\begin{figure}
    \centering
    \includegraphics[width=1\linewidth]{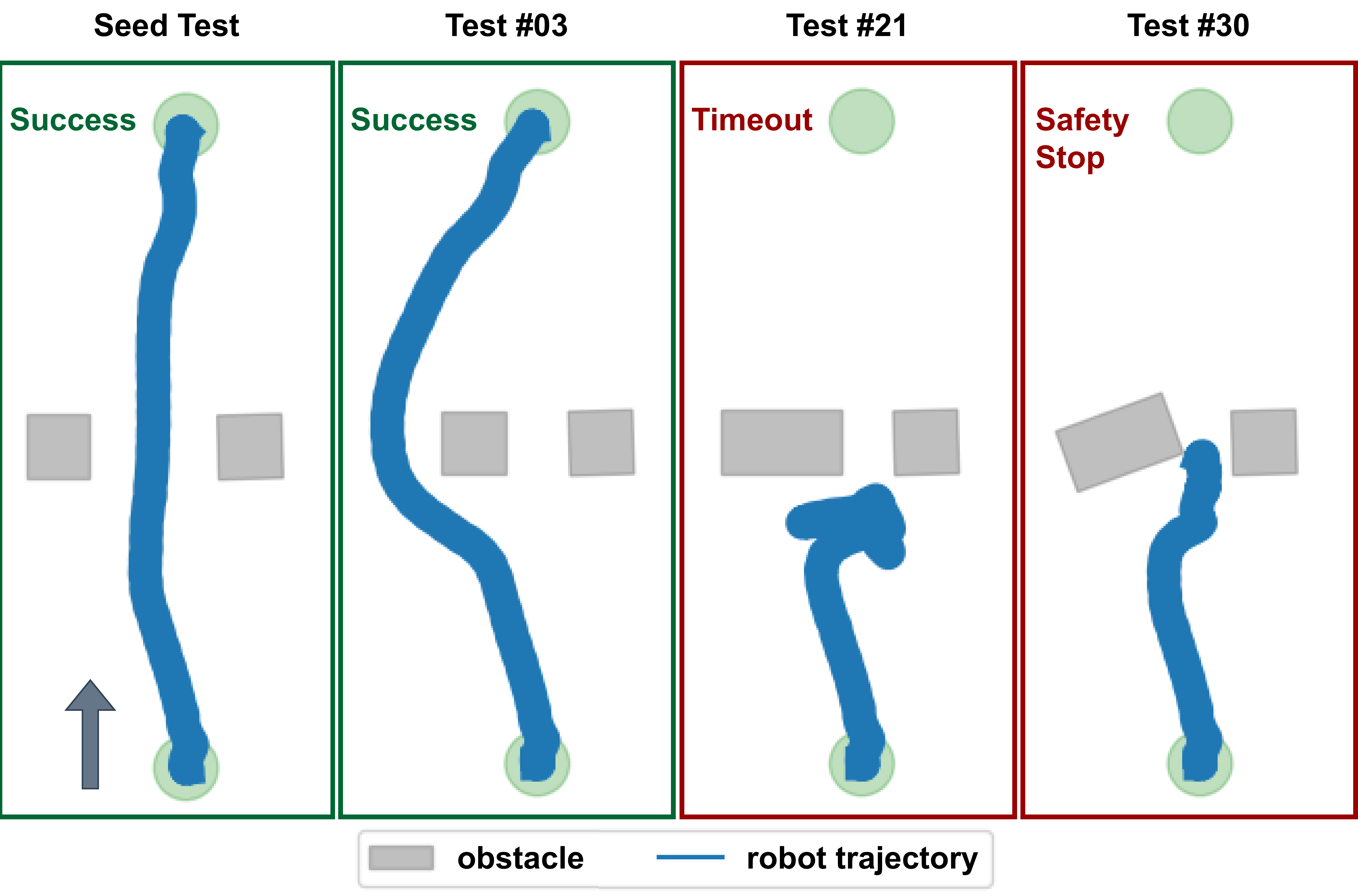}
    \caption{\small{\anybotics{Test generation process using Surrealist for a 2D robot navigation task, where the robot navigates to a target pose ($x,y$ position and $yaw$ angle). Starting with a manually defined seed test containing two obstacles (left), the system iteratively modifies the scenario by moving, resizing, and rotating the left obstacle to actively induce failures of the navigation algorithm. 
    This uncovers failure cases that are often missed by manual or purely randomized testing.}}}
    \label{fig:test-generation-process}
\vspace{-4mm}
\end{figure}

Robotic navigation testing traditionally relies on manual design and physical trials~\cite{afzal2020study,3542945,9159069}, which are costly, time-consuming, and fail to cover diverse real-world scenarios~\cite{surrealist,ZohdinasabRGT23,BirchlerKDPP23,DBLP:journals/ese/BirchlerKBGP23}, especially unpredictable obstacle interactions~\cite{10870070}. 
Physical testing is essential but risky for edge cases, making comprehensive scenario coverage and robust validation difficult.
Simulation-based testing offers a flexible, effective alternative to traditional methods~\cite{afzal2018crashing,afzal2021simulation,birchler2024roadmapsimulationbasedtestingautonomous}. High-fidelity simulators accurately model robots, sensors, and environments, enabling safe, repeatable, and cost-effective testing of navigation and control. They also facilitate systematic exploration of the operational design domain, including rare or hazardous edge cases~\cite{elbaum2021fuzzing,elbaum2021world,surrealist}.

In prior work on aerial robotics~\cite{aerialist,surrealist,superialist,SBFT-UAV2024,SBFT-UAV2025,ICST-UAV2025}, we presented two frameworks: \textit{Surrealist}\cite{surrealist}, a search-based tool that uses evolutionary algorithms to generate realistic UAV simulation tests, and \textit{Aerialist}\cite{aerialist}, a PX4-based UAV test bench. \textit{Surrealist} uncovered critical PX4~\cite{px4-Autpilot,KhatiriSZVPP24} failure cases involving obstacle avoidance.
Here, we extend them to a state-of-the-art quadrupedal robot to evaluate the benefits of automated test generation in an industrial workflow.

\begin{figure*}[]
    \centering
    \begin{minipage}[b]{0.38\textwidth}
        \centering
        \includegraphics[width=\linewidth]{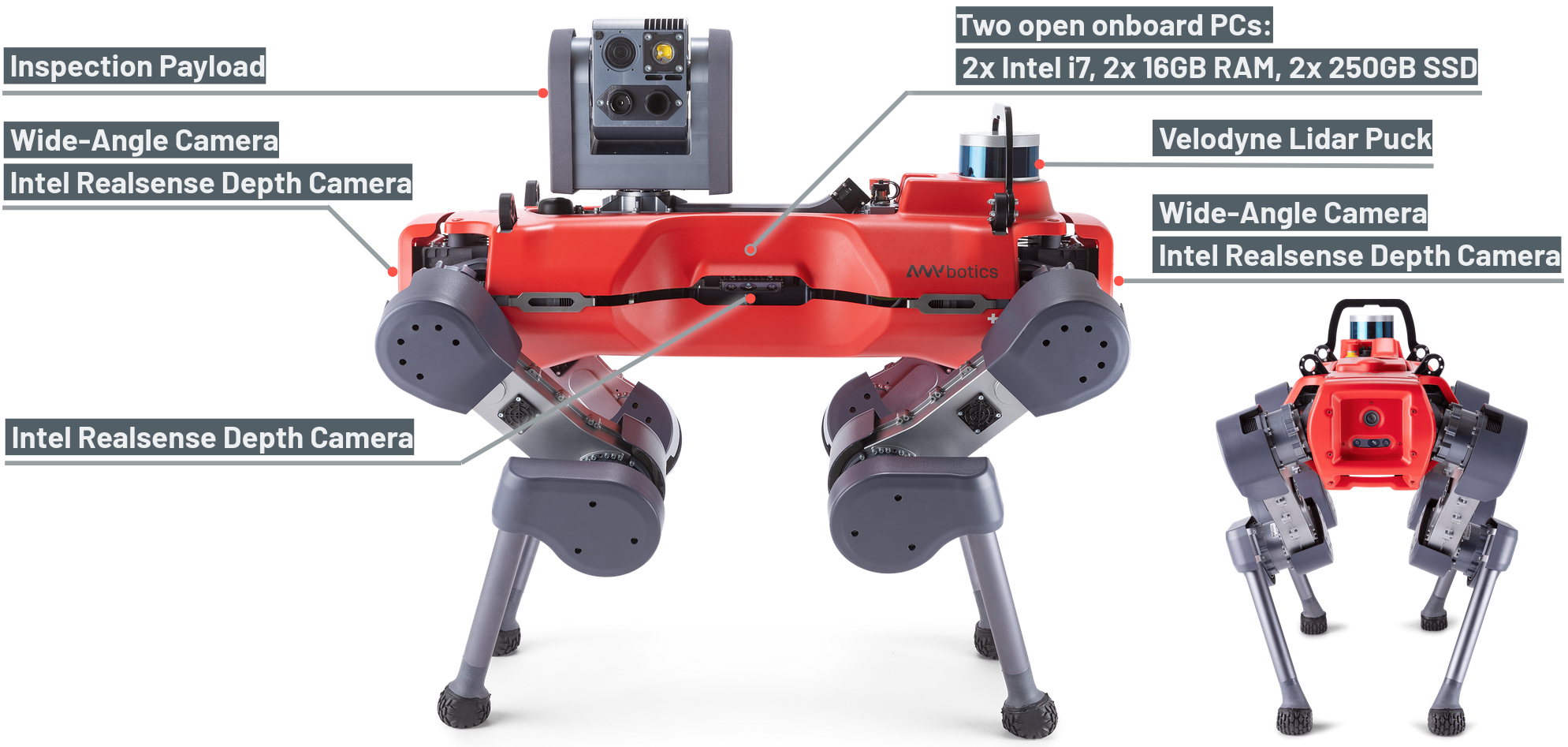}
        \caption{\small{\anybotics{ANYmal D architecture and sensors~\cite{anybotics_website}}}}
        \label{fig:anymal}
    \end{minipage}
    \hfill 
    \begin{minipage}[b]{0.57\textwidth}
        \centering
        \includegraphics[width=\linewidth]{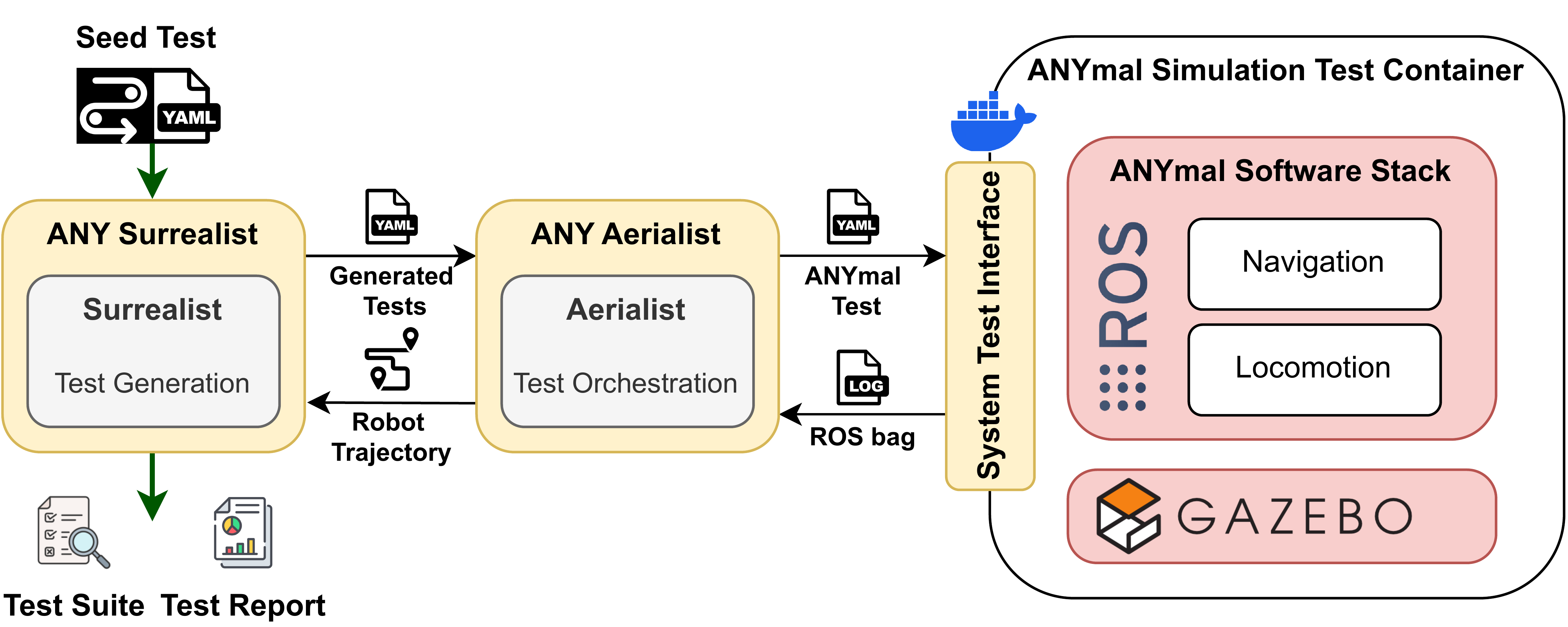}
        \caption{\small{\anybotics{Architecture of our simulation-based test generation framework}}}
        \label{fig:architecture}
    \end{minipage}
    \vspace{-4mm}
\end{figure*}

This study was conducted in close collaboration with \textit{ANYbotics}\footnote{https://www.anybotics.com/}, a world leader in developing quadrupedal robots for autonomous inspection and monitoring in industrial environments. The evaluation used their flagship platform \textit{ANYmal D}~\cite{hutter2017anymal}, shown in Figure \ref{fig:anymal}. The integration of our tools consisted of a pilot and deployment phase. During the pilot, we adapted the \textit{Surrealist}~\cite{surrealist} and \textit{Aerialist}~\cite{aerialist} frameworks to test quadrupedal navigation algorithms (see Figure~\ref{fig:architecture}). The adapted approach was initially evaluated using two experimental navigation algorithms (\textit{Exp-Nav-A} and \textit{Exp-Nav-B}) as test subjects and proved highly effective; for instance, test suites automatically generated in five scenarios revealed critical weaknesses in \textit{Exp-Nav-A}, leading to a mission success rate of just 40.3\%. These challenging test suites also provided an effective benchmark, demonstrating the superior robustness of \textit{Exp-Nav-B}, which achieved a 71.2\% success rate under the same conditions. Figure~\ref{fig:test-generation-process} illustrates examples from the test generation process for \textit{Exp-Nav-B}.

After a successful pilot, the \textit{ANYbotics} team integrated the framework into their development workflow. Over a period of six months, they first generated three test suites for their current released version (\textit{ANY-NAV-A}), assessed its performance, and identified its core deficiencies to improve in the next version. The same test suites were used to benchmark four internal candidates for the next release (\textit{ANY-NAV-B$_{1-4}$}), allowing them to effectively compare algorithms and iterate faster.  
While performance details remain confidential, feedback from a formal survey completed by eight engineers confirmed the framework’s effectiveness.
Specifically, engineers reported that ``the framework significantly streamlined their workflow by automating challenging test generation, proved effective at uncovering algorithm deficiencies missed by manual methods, and enhanced their ability to objectively benchmark algorithm improvements.''
While the evaluation centered on the ANYmal quadruped, the methodology is generalizable to other mobile robots.

The main contributions of this paper are:
\begin{itemize}[leftmargin=*, topsep=0pt, itemsep=0pt, parsep=0pt]
    \item Extension of the UAV-focused test generation framework (\textit{Surrealist}\cite{surrealist}, \textit{Aerialist}\cite{aerialist}) to quadruped navigation—introducing challenges from high-dimensional locomotion, terrain contacts, and tightly coupled perception and planning in cluttered environments; 
    
    \item Industrial evaluation at \textit{ANYbotics AG} using the \textit{ANYmal} quadruped over six months enabled engineers to efficiently: A) identify algorithm failures, B) iterate faster, and C) benchmark consistently; 

    \item Empirical insights on challenges and trade-offs in simulation-based validation of industrial robotics, from a survey involving eight ANYbotics engineers. 
\end{itemize}
\color{black}

The paper is organized as follows: Section \ref{sec:background} covers the ANYmal robot and testing frameworks. Sections \ref{sec:approach} and \ref{sec:methodology} detail our integration and methodology. Section \ref{sec:results} presents evaluation results, followed by discussion, threats, and related work in Sections \ref{sec:discussion} and \ref{sec:relatedworks}. 
We conclude and outline directions for future work in Section \ref{sec:conlusion}.

\section{Background}
\label{sec:background}

\subsection{Robotic Locomotion and Navigation}
    
    Robotic mobility—the ability to move purposefully through the environment—is vital in applications such as industrial inspection, logistics, and search and rescue \cite{Ingrand19,app13042304}. It is typically tackled via a hierarchy: \textit{locomotion} controls low-level actuators for stable movement, and \textit{navigation} plans high-level paths to goals while avoiding obstacles \cite{hoeller2024anymal}.  
    Locomotion strategies vary with robot design and use~\cite{02783649241312698}. Wheeled robots excel on smooth terrain, while tracked ones provide better mobility on uneven surfaces~\cite{solmaz2024robust}. Aerial robots maneuver in 3D~\cite{solmaz2024robust} and legged robots, especially quadrupeds, navigate stairs and rough terrain with diverse gaits like walking and trotting~\cite{hoeller2024anymal,miki2022learning,yue2020learning}. Stable, efficient movement relies on advanced control methods such as Model Predictive Control~\cite{liu2024deployment} and reinforcement learning\cite{yue2020learning,yan2024advanced,hoeller2024anymal}.

   Complimentary, navigation addresses the intelligent decision-making required for a mobile robot to reach a desired destination safely and efficiently \cite{02783649241312698}. Common navigation scenarios range from simple point-to-point movement to complex tasks like traversing a cluttered warehouse for a wheeled robot, inspecting a multi-story building with a drone, or navigating rough terrain with a quadrupedal robot. Robust navigation relies on Perception, integrating sensor data—cameras, LiDAR, and IMUs—to build an environmental model \cite{pandey2017mobile}. 
    Navigation systems must plan optimal or near-optimal paths, avoid static and dynamic obstacles, and adapt to unforeseen environmental changes~\cite{pandey2017mobile}. Planning typically involves \textit{global planning}, which maps a high-level route to the goal, and \textit{local planning}, which manages real-time obstacle avoidance~\cite{pandey2017mobile}. For quadrupedal robots, navigation is closely tied to locomotion control, as gait and dynamic capabilities influence terrain traversal and obstacle avoidance~\cite{dudzik2020robust}.

    With the rise of commercially available quadrupedal robots like ANYmal~\cite{hutter2017anymal}, Spot~\cite{raibert2008bigdog}, Unitree Go~\cite{unitree_go2}, and Mini-Cheetah~\cite{katz2019mini}, research on navigation using these platforms has surged. Recent work spans reinforcement learning for terrain adaptation~\cite{yan2024advanced}, dynamic maneuvers like obstacle jumping~\cite{gilroy2021autonomous}, and robust navigation via hierarchical state estimation~\cite{dudzik2020robust}. Efforts also address safe operation in dynamic settings using control barrier functions~\cite{dai2024sailing}, anisotropy-aware planning~\cite{zhang2024agile}, and combining locomotion with high-level perception for complex tasks like parkour~\cite{miki2022learning}. Human-robot safety is also being enhanced through visual tracking and predictive control~\cite{karlsson2022ensuring}. These advances would have been impossible without systematic simulation-based testing, which remains essential for ensuring robustness and reliability.

\subsection{The ANYmal Quadrupedal Robot} 
    The \textit{ANYmal} robot, developed by \textit{ANYbotics}, is a quadrupedal platform built for autonomous inspection and monitoring in complex industrial environments \cite{hutter2017anymal}. Tasks typically include navigating to designated locations, collecting sensor data (e.g., visual, thermal, gas), and detecting anomalies such as leaks, corrosion, or equipment faults \cite{gehring2021anymal}. Its legged design enables mobility over human-centric terrains like stairs.
    
    Figure~\ref{fig:anymal} shows  \textit{ANYmal D}, the latest commercial version, with key components and sensors. It includes a 360° LiDAR for mapping and obstacle detection, wide-angle front and rear cameras for navigation, and six depth cameras for all-around terrain perception~\cite{anybotics_website}. These sensors provide rich data for robust obstacle avoidance. In this study, ANYmal represents advanced quadrupedal robots with dynamic locomotion, multi-modal perception, and autonomous navigation~\cite{hutter2017anymal}.

\color{black}

\subsection{Surrealist and Aerialist Frameworks}
    Surrealist\footnote{https://github.com/skhatiri/Surrealist}~\cite{surrealist} is a simulation-based test generation approach for UAVs that uses real flight logs to enhance the realism and effectiveness of testing. It operates in two phases: \emph{replication}, which reconstructs real flights by optimizing obstacle configurations in simulation, and \emph{generation}, which explores variations of the replicated environment to generate more challenging scenarios. These variations, primarily obstacle modifications, help reveal weaknesses or unsafe behaviors by pushing the system close to real-world limits. Surrealist leverages an evolutionary search-based algorithm to iteratively find obstacle properties with the best 'fitness' (detailed in \cite{surrealist}).

    Built on the Aerialist\footnote{https://github.com/skhatiri/Aerialist}~\cite{aerialist} test bench, Surrealist inherits capabilities for defining test cases, generating simulation configs, parallel test execution (via Kubernetes), and automated result analysis for PX4 firmware~\cite{px4-Autpilot}. 
    The versatility of Aerialist has established it as a foundational component in UAV test generation research~\cite{SBFT-UAV2024,SBFT-UAV2025,ICST-UAV2025}.
    Building on the success in the UAV domain, this paper adapts the combined search-based test generation framework to quadruped navigation, showing cross-domain applicability.

\section{Integration Approach}
\label{sec:approach}
   
We propose an end-to-end simulation-based test generation framework for quadruped navigation and obstacle avoidance. 
Our methodology integrates three key components (Figure~\ref{fig:architecture}): the \textit{ANYmal} simulation test infrastructure, the \textit{ANY\,Aerialist} test bench, and the \textit{ANY\,Surrealist} test generator. The aim is to automatically generate challenging simulation-based test scenarios (including tricky obstacle configurations) to  test ANYmal's autonomous navigation requirements, e.g., safety and reliability.  
\textit{ANY\,Surrealist} generates test scenarios, \textit{ANY\,Aerialist} executes them via a new ANYmal interface, and the results are analyzed for potential weaknesses. Yellow highlights in Figure~\ref{fig:architecture} indicate integration extensions, detailed below.

\begin{figure*}
    \centering
    \includegraphics[width=\linewidth]{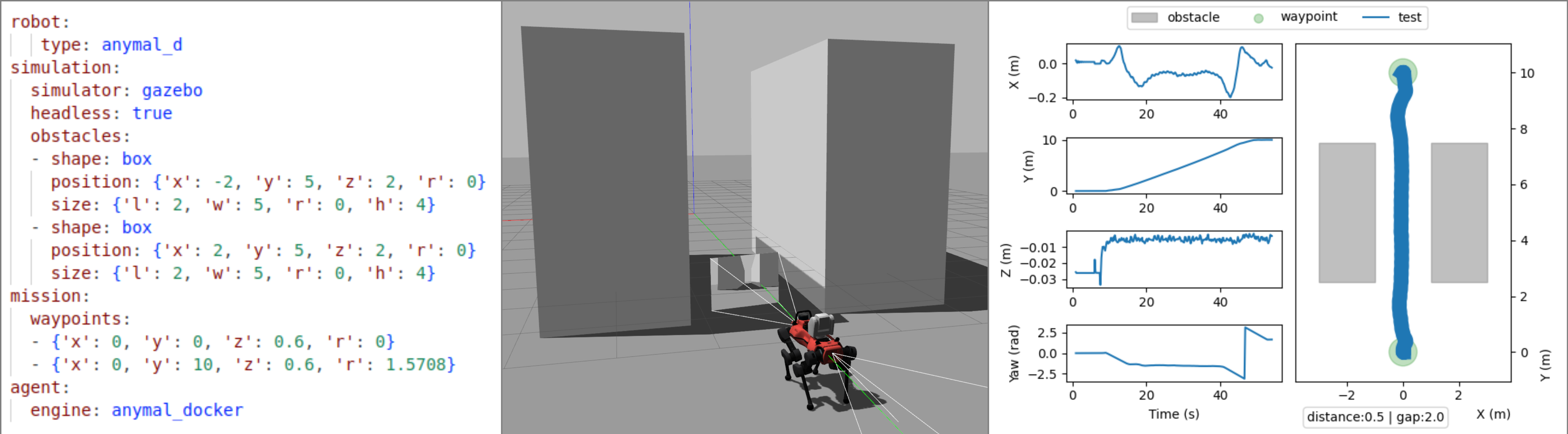}
    \caption{\small{\anybotics{Sample ANYmal test case: description (left), simulation environment (middle), robot trajectory (right)}}}
    \label{fig:aerialist_sample}
    \vspace{-4mm}
\end{figure*}

\subsection{System Under Test (ANYmal) Adaptations}
    \label{sec:approach-interface}
    \textit{ANYmal}, like other complex robots, has a large codebase, advanced software architecture, and a sophisticated multi-computer deployment setup. 
    To simplify integrating open-source test generation tools (Aerialist and Surrealist) with ANYmal and similar robots, we created the \textit{System Test Interface}—a facade that abstracts software setup, simulation configuration, and test execution using our previously defined \textit{test definition}~\cite{surrealist, aerialist}, which specifies the system, simulation, mission, and runtime commands.

    To apply this to ANYmal, we built on its existing testing setup. ANYbotics provides a robust simulation-based infrastructure using the Gazebo simulator and a detailed ANYmal D model with full kinematics, dynamics, and sensors. We added a command line interface (CLI) to support external, automated scenario configuration and control.
    The core is an Aerialist test definition file (Figure \ref{fig:aerialist_sample}, left), enabling precise specification of: 
        \begin{itemize}[leftmargin=*, topsep=0pt, itemsep=0pt, parsep=0pt]
        \item \textit{Obstacle Parameters:} number, size, shape, and position of static obstacles within the simulated environments
        \item \textit{Mission Plan:} the robot's designated starting position, goal location, and any intermediate waypoints
        \item \textit{Robot Configuration:} key parameters relevant to the specific robot setup being tested
        \end{itemize}
    
    This standardized configuration enables automated scenario generation by external tools (e.g., Surrealist) without modifying the ANYmal code base. As shown in Figure \ref{fig:architecture}, the interface bridges internal software and simulation dependencies, translating configuration parameters into commands for the ANYmal stack and Gazebo. To ensure portability and integration, the entire system—including the simulator, ANYmal software, and interface—is dockerized.

\subsection{Aerialist Adaptations (ANY\,Aerialist)}
\label{sec:approach-aerialist}

    Originally developed for UAV testing, Aerialist~\cite{aerialist} was closely tied to PX4’s features and simulation. Yet, its core idea—abstracting dependencies for black-box testing—is widely applicable. To extend Aerialist to other use cases like ANYmal, it was refactored to decouple the generic test definition, orchestration, and analysis from the use-case-specific test execution. This transformation evolved Aerialist from a PX4/UAV-specific tool into an extensible test bench for robotic navigation systems. Key modifications include:

        \begin{itemize}[leftmargin=*, topsep=0pt, itemsep=0pt, parsep=0pt]
        \item \textit{Refactoring for Extensibility:} The core Aerialist code was refactored for better modularity by introducing abstract interfaces for robot- and simulator-specific functions.
        \item \textit{External Test Execution:} Aerialist was updated to enable external test execution through the ANYmal's system test interface (previously limited to PX4 commands). 
        \item \textit{ROS Bag Support:} Aerialist was modified to read ROS bag files—the standard format in ROS (Robot Operating System) used by ANYmal—instead of PX4 ulog files.
        \item \anybotics{\textit{Failure Categories:} We automatically distinguish common categories of test outcome: \textit{Success} (the robot reaches the navigation goal within a tolerated position error, in the given time, and without colliding against obstacles), 
        \textit{Safety-Stop} (a safety function prevents the robot to move further when it gets critically close to obstacles), \textit{Timeout} (the robot can not reach the goal in time). }
        \item \textit{Geometric Calculations:} Distance functions were updated to model ANYmal as a box, reflecting its true size and shape, rather than treating it as a point-mass UAV.
        \item \textit{Enhanced Plotting:} 
        Visualization improvements include thick trajectory lines showing the robot's size and data on minimum robot-to-obstacle \textit{distance} and obstacle-to-obstacle \textit{gap} (Figure \ref{fig:aerialist_sample}, right). This tool was key for the ANYbotics team for quickly diagnosing test failures and identifying common failure patterns. 
        \end{itemize}

\subsection{Surrealist Adaptations (ANY\,Surrealist)}
\label{sec:approach-surrealist}
    While Surrealist supports two use cases, we focus only on test generation, leaving real-world test replication for future work. 
    Since Surrealist’s dependency on PX4 are abstracted via Aerialist, the above adaptations to Aerialist were enough to enable test case generation for ANYmal by the existing version of Surrealist. However, Surrealist was further enhanced for quadrupedal robot navigation, based on initial feedback from the ANYbotics team. Key adaptations include:

        \begin{itemize}[leftmargin=*, topsep=0pt, itemsep=0pt, parsep=0pt]
        \item 
        \textit{Fitness Function:} We adapted the original Surrealist fitness function, aiming to minimize the robot’s minimum \textit{distance} to obstacles over test generation iterations. According to developers, this is still a good proxy for the new context, since the likelihood of failures (i.e., collisions) increases if the robot navigates too close to obstacles.
        \item \textit{Seed Solutions:} Surrealist generates test suites based on a given \textit{seed test}. The algorithm manipulates the size and position of the obstacles in the environment to find challenging test cases based on the above fitness function. 
        We manually crafted seed tests reflecting common and challenging industrial inspection tasks (e.g., narrow corridors, doorways, cluttered spaces), based on ANYbotics' input. 
        \item \textit{Waypoint Mutation:} For certain test scenarios (e.g., robot passing a standard doorway), it is desirable to manipulate mission waypoints (i.e., the starting position) instead of the obstacles to find corner cases, as the robot may detect the same obstacles from different viewpoints. We implemented new mutations to move the waypoints in the environment, addressing such requests. 
        \item \textit{Test Suite Re-execution:} Added support for rerunning existing test scenarios to enable regression testing across navigation algorithm versions.
        \item \textit{Performance Metrics:} Extended logging to capture performance and quadrupedal-specific metrics per test, including minimum obstacle distance, gap between obstacles, path lengths, deviation, and test duration.
        \item \textit{Metrics Aggregation:} Aggregate logs summarize test outcomes (success, safety-stop, timeout) with descriptive statistics (min, max, average) of the above performance metrics. 
        \end{itemize}

\subsection{Development Process and Integration Workflow}
We forked the public Surrealist and Aerialist repositories for ANYmal-specific adaptations, refactoring the core frameworks to improve modularity while keeping the forks synchronized for easy maintenance. This design proved effective, as the integration required minimal code: Surrealist was extended with just three new classes extended from generic base classes ($\sim$350 LOC), and Aerialist with two ($\sim$250 LOC). These minimal additions underscore the framework's modularity and facilitate extension to other robotic use cases, following the detailed guidelines documented in the public repositories.

The full integration—including developing the ANYmal interface, refactoring the framework, and testing—took three months, with the first author dedicating roughly 80\% of his time. 
Support from the ANYbotics team, facilitated through bi-weekly meetings and ongoing feedback, was essential to the success of this process.

\section{Research Methodology}
\label{sec:methodology}
Our empirical study evaluates our integrated simulation-based test generation framework to enhance quadrupedal robot navigation robustness, focusing on obstacle avoidance. It has two phases: (1) adapting and validating our framework on ANYbotics’ navigation software with two experimental obstacle avoidance algorithms, and (2) integrating it into ANYbotics' workflow. We use quantitative metrics and qualitative feedback, guided by four research questions:

\begin{itemize}[leftmargin=*, topsep=0pt, itemsep=0pt, parsep=0pt]
    \item \textit{RQ$_1$ [Development Process]:} How does the integrated framework improve the development workflow  and testing practices at ANYbotics?
    \item \textit{RQ$_2$ [Failure Detection]:} How effective is Surrealist at generating test cases that reveal failures in ANYmal's obstacle avoidance?
    \item \textit{RQ$_3$ [Improvement Assessment]:} Can Surrealist track and quantify performance improvement across navigation software versions?
    \item \textit{RQ$_4$ [System Verification]:} How does the integrated framework enhance the verification process for the ANYmal navigation, ensuring robustness and reliability?
\end{itemize}

\subsection{Pilot}

    In the pilot phase, we evaluated the integrated framework using two experimental obstacle avoidance algorithms from ANYbotics, treated as black-box test subjects: \textit{Exp-Nav-A} and \textit{Exp-Nav-B}. This demonstrated our approach without accessing their proprietary navigation stacks. The experiments were:

    \begin{enumerate}[leftmargin=*, topsep=0pt, itemsep=0pt, parsep=0pt]
        \item \textit{Scenario Definition (RQ$_1$):} We created five seed scenarios based on real-world use cases like narrow corridors and cluttered spaces.
        \item \textit{Test Generation (RQ$_2$):} We used Surrealist to generate a test suite (TS-Exp-A) from the seed tests for the first algorithm (Exp-Nav-A) and analyzed its performance.
        \item \textit{Algorithm Comparison (RQ$_3$):} We executed the same test suite (TS-Exp-A) on a newer algorithm (Exp-Nav-B) to compare their performance.
        \item \textit{Test Suite Comparison (RQ$_2$/RQ$_3$):} We generated a new test suite (TS-Exp-B) tailored to Exp-Nav-B to assess how algorithm differences affect test generation.
    \end{enumerate}

    During the three-month pilot, the team included the first author as lead developer and three ANYbotics supervisors (navigation and test engineers). Iterative feedback enabled continuous improvements.  
    We evaluated the obstacle avoidance performance in the generated scenarios using the metrics introduced in Section \ref{sec:approach-surrealist} including the Mission Success Rate and Safety-Stop Rate.
    The test generation process was conducted by the first author on a development laptop equipped with \textit{Intel Core Ultra 9 185H} CPU, \textit{NVIDIA RTX 2000 Ada} GPU, and \textit{64 GB} of RAM.
    The first author analyzed each algorithm's overall performance, identified weaknesses and behavioral differences, and reported findings to the development team.
    The pilot phase primarily addresses RQ$_{1,2}$ [Development Process, Failure Detection], while providing preliminary insights for RQ$_3$ [Improvement Assessment].

\subsection{Deployment}
    Following the successful pilot, we deployed the framework at ANYbotics, where the navigation team integrated it to evaluate in-house obstacle avoidance algorithms. A senior engineer, not involved in the pilot, led its use for failure identification, analysis, performance comparison, and improvement guidance. The first author provided technical support, while four other navigation team members participated in the discussions. 
    In the first six months post-pilot, we tested multiple versions of the proprietary ANYmal navigation stack, including the current release referred to as \textit{ANY-Nav-A}, and four internal candidates for the next release, referred to as \textit{ANY-Nav-B$_{1-4}$}:

  \begin{enumerate}[leftmargin=*, topsep=0pt, itemsep=0pt, parsep=0pt]
    \item \textit{Test Generation (RQ$_2$):} The ANYbotics team used Surrealist to generate a test suite (TS-ANY-A) for ANY-Nav-A, based on a selection of 3 out of 5 pilot study seed scenarios. They assessed its performance and identified the deficiencies to improve in the next version. 
    \item \textit{Algorithm Comparison (RQ$_3$):} During their development workflow for the next version, they used the tests in TS-ANY-A as a benchmark. They executed the same tests with ANY-Nav-B$_{1-4}$ as test subjects to assess performance improvements and behavioral differences. 
    \item \textit{Targeted Test Generation (RQ$_2$ \& RQ$_4$):} During ANY-Nav-A field tests, engineers found two new failure types. The development team asked the first author to create Surrealist test suites targeting these failures for simulation-based diagnosis and resolution. 
\end{enumerate}

Since the quantitative results from the six-month deployment are confidential (including the generated test suites, the identified failure cases, and the performance of proprietary algorithms), we evaluated the framework's industrial impact primarily through a formal questionnaire. To provide context, participants from various teams (navigation, locomotion, perception, and verification) first watched a 14-minute video demonstrating the project's workflow. 
After establishing participant context with demographics, the survey assessed the usability of the end-to-end workflow, from installation and test definition with Aerialist to automated generation with Surrealist. It then evaluated the framework's effectiveness based on the test suite quality, benchmarking performance, and targeted debugging. Finally, the survey assessed the system's overall impact and future directions.
While some details are confidential, general findings from this survey are discussed in Section \ref{sec:results}, contributing to RQ$_{1-4}$ 
[Development Process, Failure Detection, Improvement Assessment, System Verification] 
with an emphasis on real-world adoption and impact.

\color{black}

\section{Evaluation Results}
\label{sec:results}
\begin{table*}[]
    \centering
    \caption{Performance comparison of the pilot test subjects.}
    \label{tab:pilot_performance}
    \begin{tabular}{ll rrrrrrrrrrrr}
        \toprule
        \multirow{2}{*}{Test Subject} & \multirow{2}{*}{Test Suite (\#)} & \multicolumn{2}{c}{boxes$_1$} & \multicolumn{2}{c}{boxes$_2$} & \multicolumn{2}{c}{corridor} & \multicolumn{2}{c}{cylinders} & \multicolumn{2}{c}{L-corridor} & \multicolumn{2}{c}{overall} \\
        \cmidrule(lr){3-4} \cmidrule(lr){5-6} \cmidrule(lr){7-8} \cmidrule(lr){9-10} \cmidrule(lr){11-12} \cmidrule(lr){13-14}
         & & {Succ.} & {S-Stop} & {Succ.} & {S-Stop} & {Succ.} & {S-Stop} & {Succ.} & {S-Stop} & {Succ.} & {S-Stop} & {Succ.} & {S-Stop} \\
        \midrule
        Exp-Nav-A & TS-Exp-A (429) & 39.2\% & 40.5\% & 42.1\% & 44.9\% & 37.5\% & 39.8\% & 75.0\% & 23.2\% & 23.1\% & 52.9\% & 40.3\% & 42.2\% \\
        Exp-Nav-B & TS-Exp-A (429) & 81.1\% & 5.4\%  & 69.2\% & 0.0\%  & 75.0\% & 12.5\% & 98.3\% & 1.8\%  & 48.1\% & 16.4\% & 71.2\% & 7.7\%  \\
        \midrule
        Exp-Nav-B & TS-Exp-B (422) & 72.5\% & 12.2\% & 89.8\% & 8.3\%  & 64.7\% & 11.8\% & 76.3\% & 22.0\% & 34.8\% & 5.6\%  & 68.3\% & 11.1\% \\
        \bottomrule
    \end{tabular}
    \vspace{-4mm}
\end{table*}

This section presents the evaluation results, organized according to the four research questions outlined in Section \ref{sec:methodology}. For each question, we synthesize findings from both the pilot and deployment phases, combining quantitative metrics with qualitative feedback from the ANYbotics team. The quantitative analysis primarily uses metrics from the pilot study to objectively address RQ$_{2,3}$. The qualitative insights, drawn from iterative discussions and a formal survey with engineers, provide the industrial perspective on RQ$_{1-4}$.

A total of 8 engineers from ANYbotics participated in the feedback survey, representing a diverse range of relevant teams: Navigation (including 3 navigation engineers, a team lead, and a product manager), Locomotion (1), Perception (1), and Verification (1). Participants were highly experienced and educated; 7 participants have more than three years of professional robotics experience, and all hold a master's degree (5) or PhD (3) in either Robotics or Electrical Engineering.
Their self-reported expertise is strong, with most participants rating themselves as 'proficient' or higher in robotic navigation, general robotic testing, and simulation-based testing, including a test automation expert. Regarding the integrated framework itself, 3 participants had direct `hands-on experience' or were a `regular user', while the remaining 5 were `conceptually familiar' through presentations and discussions.

\subsection{RQ$_1$ [Development Process]}
    \subsubsection{Pilot} 
    The pilot phase used an iterative process guided by continuous feedback from the ANYbotics team via informal discussions and demos. This collaboration was key to improving usability and practical value. Many features in Sections \ref{sec:approach-aerialist}, \ref{sec:approach-surrealist} were added based on their input: improved plotting, clearer test outcome distinctions, and scenario selection in Aerialist, as well as test suite re-runs, waypoint mutation, and detailed metric collection in Surrealist.

    \subsubsection{Deployment} 
    In the deployment phase, feedback from the formal questionnaire (Likert scale: 1 = very low, 5 = very high) confirmed the system’s overall usability and its positive impact on development workflows. Setup documentation received a high average rating of 4.1, although the installation process was perceived as moderately complex (average = 3). The primary challenge was coordinating the Docker containers, as it \textit{``still requires a few manual steps to get it working"}, suggesting the need for  \textit{``a bash/python script that installs everything."}

    The YAML-based test definition with \textit{Aerialist} was rated highly for usability (average = 4.5), with 75\% of participants (6 out of 8) finding it \textit{`easier'} or \textit{`much easier'} to use than their previous approaches. Test execution was similarly well received (average = 4.6). The built-in automated visualizations were also praised (average = 4.3), with around 60\% of participants rating them as \textit{`better'} or \textit{`much better'} than alternative tools. Suggestions for enhancement included more interactive and detailed features to align with tools like Foxglove\footnote{https://foxglove.dev/}: \textit{``adding the speed over time (change in color?) [...] grid on the plots [...] and an option for interactive plot would be super cool"}. 

    Regarding automated test generation with \textit{Surrealist}, defining a seed scenario was rated as moderately easy (average = 3.5), while all participants considered generating a test suite from it highly intuitive (average 4.1). In terms of efficiency, 37\% of participants reported a \textit{`significant improvement'} over existing methods, while 50\% considered it \textit{`comparable, with the main benefit being automation rather than speed'}.

    \begin{tcolorbox}[colback=gray!15!white,colframe=black, boxsep=2pt,left=3pt,right=3pt,top=3pt,bottom=3pt]
        \textbf{Finding 1}: 
        The integrated framework streamlined testing workflows, with Aerialist rated highly for usability and visualization (75\% preferred it over prior tools) and Surrealist praised for intuitive test generation. Developers reported reduced manual effort and a faster and more productive test creation. 
    \end{tcolorbox}

\subsection{RQ$_2$ [Failure Detection]}

    \subsubsection{Pilot Study}
    During the pilot, we evaluated Surrealist's ability to generate failure-inducing scenarios for two experimental navigation algorithms developed at ANYbotics. We measured each algorithm’s \textit{Success Rate}—the percentage of tests where the robot reached its goal safely—and \textit{Safety Stop (S-Stop) Rate}—the percentage of failures caused by the robot's collision safety layer, which halts the robot when its current velocity is predicted to cause a collision with nearby obstacles.

\begin{figure}
        \centering
        \includegraphics[width=\linewidth]{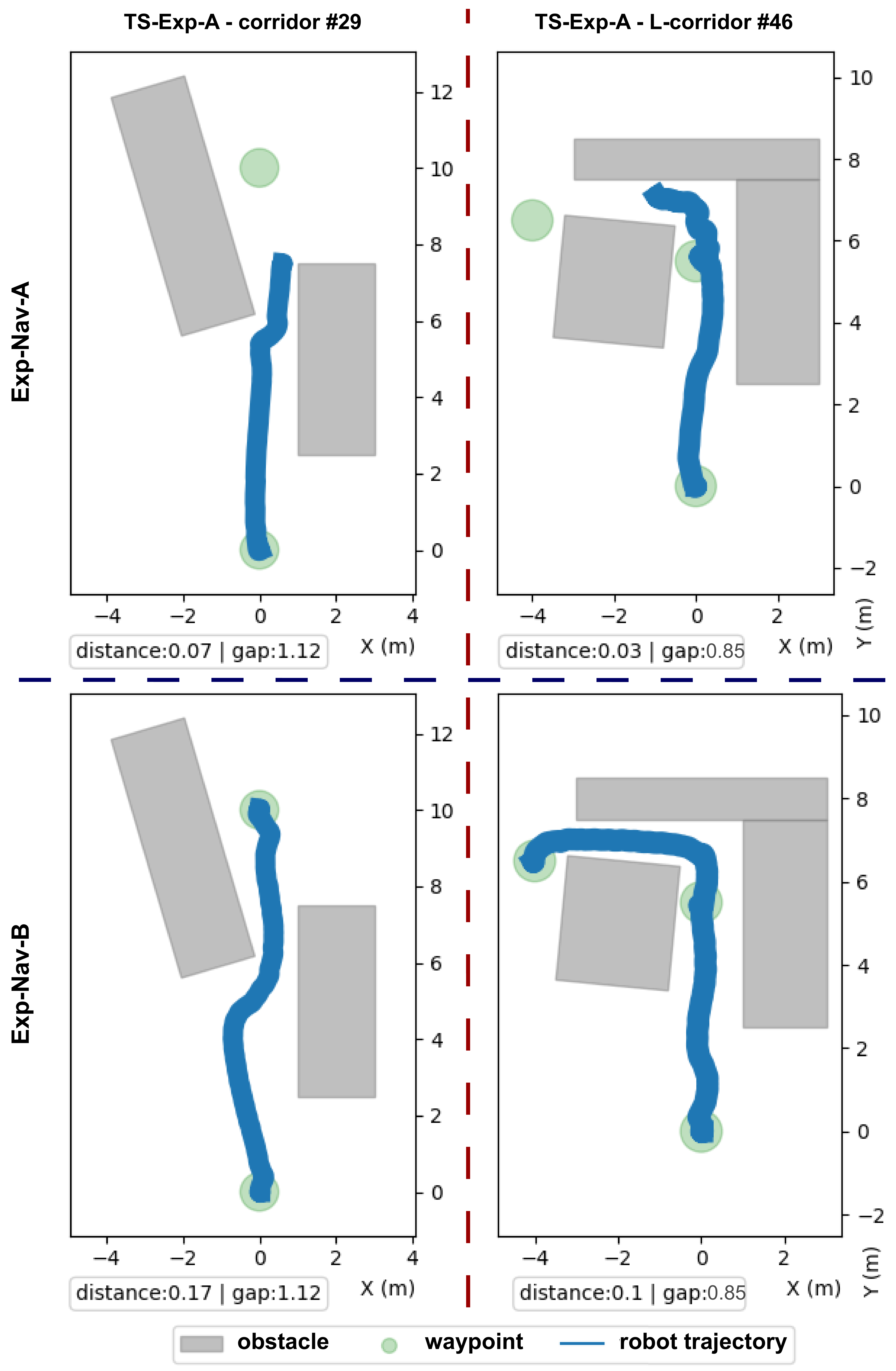}
        \caption{\small{\anybotics{Robot trajectory comparison with different obstacle avoidance algorithms in the same test environments}}}
        \label{fig:pilot_comparison}
    \vspace{-4mm}
    \end{figure}

    We first selected five representative navigation scenarios (seeds) reflecting common and challenging industrial environments. A simple 10m task from $(x, y) = (0,0)$ to $(0,10)$ was manually designed by the first author, featuring five obstacle configurations. Symmetric setups included the \textit{boxes$_1$} scenario with two 1$\times$1m boxes (Figure \ref{fig:test-generation-process}-left), two 1m-diameter \textit{cylinders}, and a narrow \textit{corridor} formed by two 5m walls (seed for Figure \ref{fig:pilot_comparison}-left), all spaced 2m apart. Asymmetric setups included \textit{boxes$_2$} with two offset boxes, and an \textit{L-corridor} (seed for Figure \ref{fig:pilot_comparison}-right). We used these five seeds to generate the test suite \textit{TS-Exp-A}, tailored for the \textit{Exp-Nav-A} algorithm.
    
    As shown in Table~\ref{tab:pilot_performance} (first row), the generated suite, which included 429 test cases in total, effectively exposed failures in Exp-Nav-A, which had a low overall success rate of 40.3\% and S-Stops accounted for 42.2\% of outcomes. Figure~\ref{fig:pilot_comparison} (top) illustrates test cases and robot behavior in two scenarios. The algorithm performed well with wider obstacle gaps but struggled significantly when gaps were $<1.5\,m$, either behaving risky and leading to a safety-stop, or behaving too cautions and not entering wide enough gaps, leading to a timeout while looking for other ways around. The \textit{L-corridor} scenario was especially difficult, with only a 23.1\% success rate and 52.9\% S-Stops, highlighting issues navigating tight spaces safely.

    \subsubsection{Deployment}

    While specific quantitative performance data from the deployment phase remains confidential, feedback from the formal questionnaire confirmed Surrealist's effectiveness in uncovering failures. Participants rated (with Likert scale: 1 = very low, 5 = very high) the generated test cases as highly realistic and relevant, with an average score of 4.3. Most participants reported marginal (50\%) or no concern (50\%) about the sim-to-real gap, with an average rating of 4.5 (where 5 means not at all concerned). The scenarios were perceived as challenging, receiving an average rating of 3.6. Notably, 50\% of participants found these scenarios more challenging than their typical manual test cases. Participants found the scenarios effective in exposing realistic failures, noting they matched those seen on real robots: \textit{``I never saw a failure case that we did not yet see with Surrealist"}. Scenarios with ``narrowing gaps" (similar to the corridors in Fig. \ref{fig:pilot_comparison}) were found specially realistic and representative of challenging industrial environments. The diversity of generated tests was appreciated, even if some were oversimplified, as the goal is to \textit{``break the system in test"}. Concerns about the sim-to-real gap centered on perception noise, dynamic elements, and oversimplified geometry: \textit{``obstacles in industrial settings are rarely as uniform as the cubes or cylinders used in testing"}. While the gap between simulation and reality was seen as manageable, especially for navigation, one participant still recommended real-world verification: \textit{``I would still retest... to verify the behavior in the real world"}.
    
    Importantly, all 8 participants rated the framework as either Effective (62\%) or Very Effective (38\%) at identifying previously unknown failures and corner cases. 
    This was reinforced by qualitative feedback; for example, one engineer commented: ``Unknown edge cases. This is were surrealist shines the most!''. 
    Moreover, the use case of generating a test suite based on a known real-world failure was highly valued. Engineers gave this approach strong ratings for `relevance' (average 4.1) and `solution validation' (average 4.4), though ratings were more moderate for `root cause analysis' (average 3.3). The effort to convert a real-world issue into a simulation seed was rated as `high' by 37\% of participants, highlighting a need for future automation, e.g., by \textit{''[automatically] creating test scenarios from collected point clouds/depth data"}.

    \begin{tcolorbox}[colback=gray!15!white,colframe=black, boxsep=2pt,left=3pt,right=3pt,top=3pt,bottom=3pt]
    \textbf{Finding 2}: 
    Surrealist's search-based approach effectively generated challenging scenarios that revealed critical failures.
    In the pilot, it reduced a preliminary navigation algorithm's success rate to 40.3\%. All engineers rated it as effective for uncovering previously unknown failures and corner cases during deployment. Scenarios were considered both realistic (avg. 4.5) and challenging (avg. 3.6). The diversity of generated tests was valued positively for stress-testing the system.
    \end{tcolorbox}

\subsection{RQ$_3$ [Improvement Assessment]}

\begin{figure}
    \centering
    \includegraphics[width=1.0\linewidth]{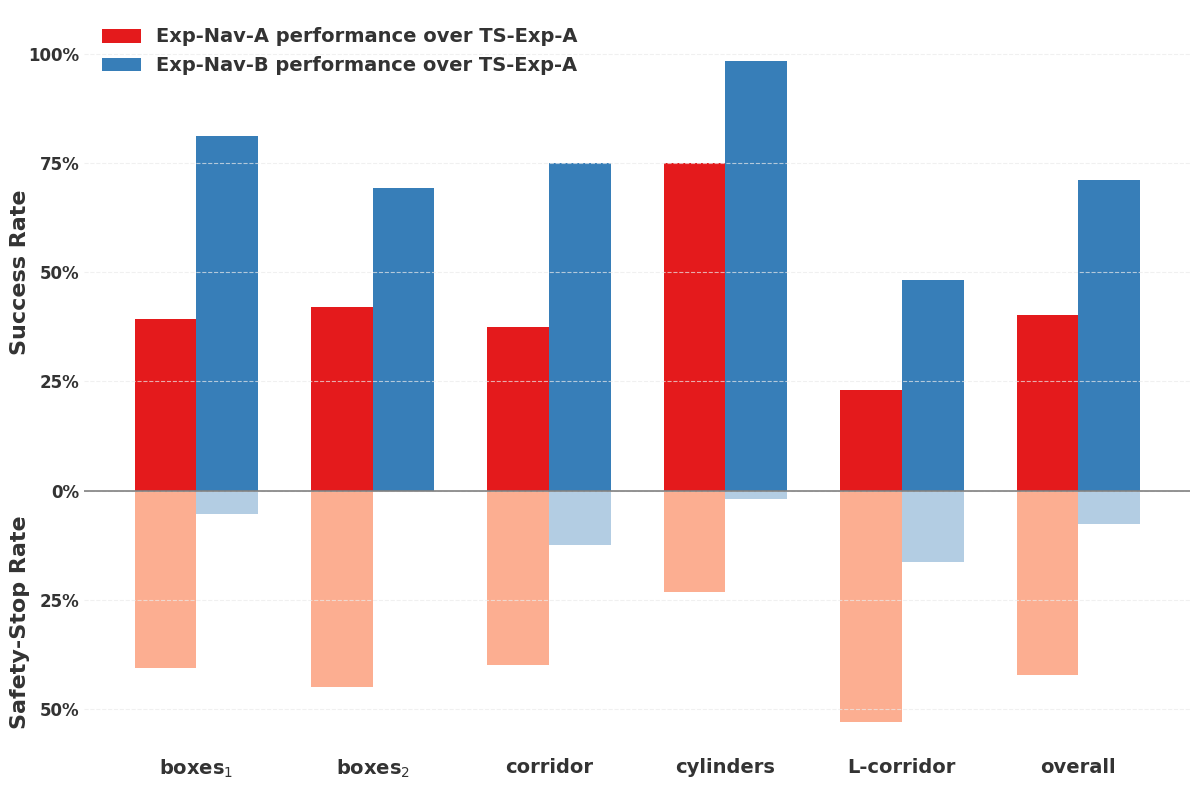}
    \caption{\small{Performance comparison of the pilot test subjects for TS-Exp-A in a two-sided bar chart. The second iteration of the obstacle avoidance algorithm showed a clear improvement (significantly higher success rate and lower safety-stop rate) over the initial prototype.}}
    \label{fig:exp_a_bar_graph}
\vspace{-4mm}
\end{figure}

\subsubsection{Pilot Study}
    A key feature we implemented during the pilot enabled the re-running of generated test suites, giving ANYbotics a repeatable, quantitative method to compare algorithms. 
    To evaluate system improvements, we tested two experimental obstacle avoidance algorithms, Exp-Nav-A and Exp-Nav-B, on the same challenging test suite, TS-Exp-A, and measured their success and S-Stop rates.
    From Table \ref{tab:pilot_performance} (rows 1-2), Exp-Nav-B outperformed Exp-Nav-A with a success rate of 71.2\% vs. 40.3\%, and a significantly lower S-Stop rate of 7.7\% vs. 42.2\%. This is even more evident in the comparison chart in Figure~\ref{fig:exp_a_bar_graph}, where Exp-Nav-B consistently shows higher success and lower s-stop rate in all test scenarios. 
    
    The qualitative nature of this improvement is visually evident in the test examples shown in Figure \ref{fig:pilot_comparison}. Across different scenarios, Exp-Nav-B consistently executes smoother, more confident paths with safer clearances from obstacles. For example, in the L-corridor \#46 scenario, Exp-Nav-B performs a wide, clean turn (distance to closest obstacle: 0.1m), whereas Exp-Nav-A takes a very tight, hesitant turn that brings it dangerously close to the corner (distance to closest obstacle: 0.03m) which causes the robot's collision safety layer to halt the robot. 
    This demonstrates that the framework not only quantifies success or failure but also provides visual evidence to assess the robot’s behavior, making it a relevant tool for validating and comparing improvements. 
    
    We then examined the framework’s capability to adapt its test generation process in order to expose the specific weaknesses of the more robust Exp-Nav-B algorithm. To this end, we generated a new test suite (TS-Exp-B) using the same seed scenarios used before, but this time optimized for the Exp-Nav-B algorithm. As shown in Table~\ref{tab:pilot_performance} (rows 2--3), the performance of Exp-Nav-B declined when evaluated on its own custom-generated test suite (422 test cases). Specifically, the overall success rate dropped from 71.2\% to 68.3\%, while the rate of safety stops increased from 7.7\% to 11.1\%. This pattern was consistent across most test scenarios, with the notable exception of \texttt{boxes$_2$}, where the success rate actually improved. However, even in that scenario, Surrealist identified new safety-critical configurations, resulting in an increase in the safety stop rate from 0\% to 8.3\%.
    These results demonstrate that our search-based framework is capable of effectively uncovering and targeting the distinct vulnerabilities of each system under test, producing a uniquely challenging test suite tailored to each individual implementation.
    
    \subsubsection{Deployment}
    The value of the framework's comparative assessment capability was strongly confirmed during the industrial deployment, where the ANYbotics team actively used the test suite re-execution feature to benchmark new implementations against the current release, and analyze their pros and cons to inform enhancement decisions.
    Feedback from the formal questionnaire highlighted the feature's high utility. All participants agreed that Surrealist can be used to compare different algorithms, rating its usefulness as higher than their previous manual comparison methods, with an average score of 4.6. Furthermore, the performance metrics provided by the system were considered `very relevant' for this purpose by most participants (7 out of 8). 
    Qualitative feedback provided suggestions for future improvements. On the analysis side, engineers suggested the development of a dedicated tool for simultaneous visualization of two test runs to make algorithm comparison less time-consuming, and the addition of confidence intervals to success rates to enhance statistical rigor. Furthermore, participants suggested new performance metrics beyond mission success, with one engineer noting it \textit{``would be nice to have some `efficiency' metric that measured how close was the robot to the `optimal'/`shortest' path to the goal and how fast it moved along it."}
    
    \begin{tcolorbox}[colback=gray!15!white,colframe=black, boxsep=2pt,left=3pt,right=3pt,top=3pt,bottom=3pt]
    \textbf{Finding 3}:
    The framework enables repeatable, quantitative comparisons of navigation algorithms across software versions. It distinguished two preliminary approaches (71.2\% vs. 40.3\% success rate, 7.7\% vs. 42.2\% safety stop rate), with qualitative visual evidence confirming behavioral improvements. During deployment, engineers rated the benchmarking feature highly (on average 4.6 out of 5), considering it a significant upgrade over previous methods and used to guide their enhancement efforts. 
    \end{tcolorbox}
    
\subsection{RQ$_4$ [System Verification]}

    \subsubsection{Pilot} The pilot study showed that Surrealist can aid system verification by generating challenging scenarios and exposing failures. Its ability to reveal issues—even with a simplified navigation stack—highlights its potential for uncovering weaknesses in more complex systems. 
    
    \subsubsection{Deployment} 
    Survey results from the deployment phase confirm that the framework adds significant value to system verification. All participants \textit{`agreed'} or \textit{`strongly agreed'} that it increased their confidence in the system's robustness, and a strong majority (7/8) said it improved their team's verification ability. 
    This was attributed to its effectiveness in specific Verification and Validation (V\&V) tasks:
    All participants rated the system as `better' or `much better' than manual design for finding difficult-to-predict corner cases, and its ability to re-run test suites for regression testing was considered `extremely valuable' by 86\% of the engineers (6 out of 7 responses). The generated scenarios and visualizations were also found to be very helpful in debugging the root cause of failures (average rating of 3.9).
    Looking at the broader workflow, a majority of participants (71\%) found the approach `very valuable' for prioritizing physical, real-world tests, and a majority (6 out of 8) see high or some potential for using the generated reports as formal V\&V evidence. The framework's critical role was summarized by one engineer who stated they ``would not release [new navigation algorithms] if they were not tested with surrealist'', positioning the tool as an essential pre-release validation gate. Complimentary, another engineer stated that ``a further validation phase in real world would be needed to cover unforeseen real world interactions/onboard resources limitations or judging qualitatively the behaviors."

    \begin{tcolorbox}[colback=gray!15!white,colframe=black, boxsep=2pt,left=3pt,right=3pt,top=3pt,bottom=3pt]
    \textbf{Finding 4}: 
    The framework enhances system verification and developer confidence by systematically uncovering difficult-to-predict corner cases and enabling efficient regression testing. Participants rated it superior to manual test design, found its debugging support highly useful, and called it an essential pre-release validation step.
    \end{tcolorbox}
    
\color{black}

\section{Discussion}
\label{sec:discussion}
\textbf{Industrial Impact \& Adoption}. 
Our research demonstrates the successful adaptation and industrial integration of the Surrealist and Aerialist frameworks into the development workflow of the ANYmal quadrupedal robot at ANYbotics. Through a two-phase evaluation, we confirmed that our approach (i) enhances the development process (RQ$_1$), (ii) effectively uncovers critical navigation failures (RQ$_2$), (iii) provides objective benchmarks for system improvement (RQ$_3$), and (iv) significantly strengthens the overall verification pipeline (RQ$_4$). 
The successful deployment of our framework for testing a proprietary navigation stack marks a key milestone, transitioning our search-based testing methodology from a research prototype to a solution demonstrated in its intended operational environment—the ANYbotics development workflow (Technology Readiness Level 7~\cite{mankins1995technology}).
\anybotics{The framework's adoption and impact are reflected in the survey results:  4 out of 8 participants identified themselves as `current users'. The impact on the team's workflow was rated as `very positive' by 6 out of 8 engineers. The majority reported they will use the tool ``occasionally for specific tasks'' (63\%) and on a monthly or weekly basis (63\%), a strong indication of sustainable integration into their development processes. 
As one engineer noted, the ability of the tool to create \textit{``test cases that reflect challenges observed later in the real world [...] built up trust quite a bit''}. 
Furthermore, the ANYbotics Locomotion team is already making efforts to ``adopt a similar search-based testing approach'',  
demonstrating how the project's impact extends beyond the initial navigation context and is shaping the company's long-term testing strategy.}

\begin{figure}
    \centering
    \includegraphics[width=1.0\linewidth]{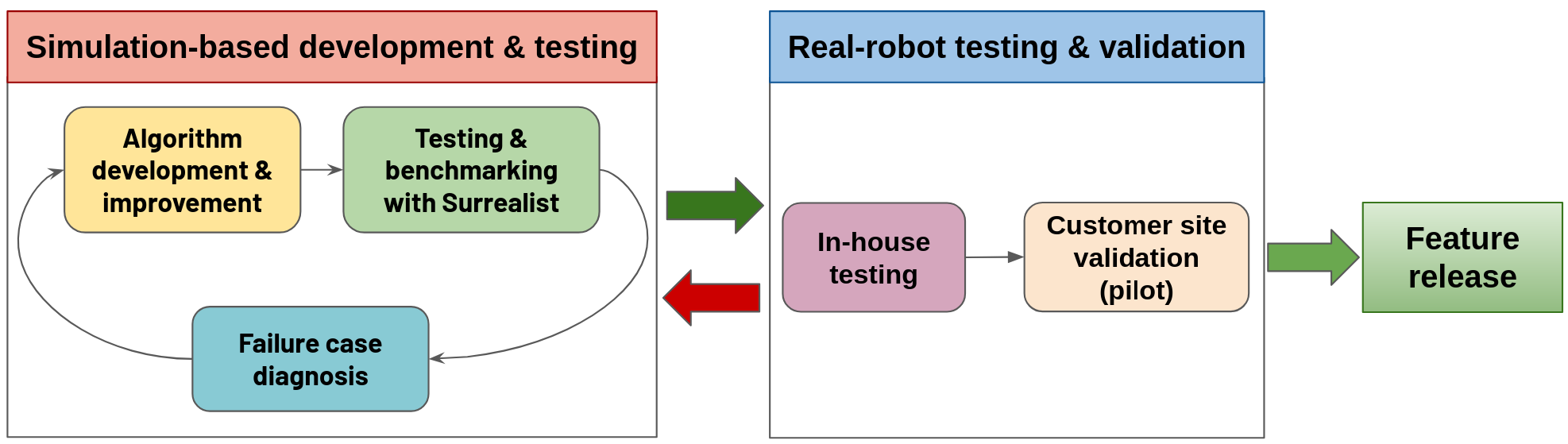}
    \caption{\small{\anybotics{ANYbotics' development workflow for releasing new navigation algorithms includes both simulation and real-robot testing stages. Our approach integrates directly into the simulation stage, allowing for early detection of critical failures and deficiencies before they reach real-robot tests. }}}
    \label{fig:dev_workflow}
    \vspace{-4mm}
\end{figure}

\textbf{Broader Applicability Across Domains \& Development Stages}.  The framework—combining automated test generation with a modular abstraction interface—has matured to a point where it can be applied beyond UAVs and quadrupedal robots. It is designed for easy extension to other domains, such as wheeled robots and manipulators, as long as an appropriate simulation interface is available. 
\anybotics{Complementary, this study offers key insights into the practical applications of automated test generation in industrial robotics. The survey provided a clear ranking of development activities, where the tool provides the most value.  
\textit{Regression Testing} emerged as the top use case selected by all participants (8/8), followed by \textit{MLOps} (6/8), highlighting the need for further automation and optimizations for a fast-paced ML development cycle. 
Other highly valued applications included \textit{Prototyping \& Early Feature Development} (5/8), \textit{Exploratory Testing} (5/8), \textit{Failure Replication \& Debugging} (4/8) and \textit{Release Validation} (4/8).}

\anybotics{
This versatility allows the framework to be deeply integrated into the ANYbotics development workflow, as illustrated in Fig. \ref{fig:dev_workflow}. ANYbotics releases new features following a two-stage process: development and testing in simulation, followed by deployment on real hardware—both at internal testing facilities and pilot customer sites. Our framework fits seamlessly into the simulation phase, enabling more thorough and efficient testing and benchmarking early in the pipeline. This early-stage integration supports the timely identification of algorithmic weaknesses, speeding up the entire development process by minimizing issues that would otherwise surface during the more costly and time-consuming real-world testing stages. Ultimately, this leads to more robust and reliable releases.}

\textbf{Feedback from Study Participants.} While the framework proved highly effective for its intended purpose, the industrial deployment and the detailed feedback from the survey also highlighted its current limitations as well as a clear path for future evolution. The suggestions centered on three key areas: 
\begin{enumerate} [leftmargin=*, topsep=0pt, itemsep=0pt, parsep=0pt]
\item \textit{Further bridge the sim-to-real gap}: add more realistic sensor noise models, generate complex 3D terrains (such as stairs and ramps), incorporate a broader range of irregular obstacles (such as those found in industrial settings), and add support for dynamic obstacles. 
\item \textit{Streamline the debugging process}: introduce a ``real-to-sim'' capability to auto-generate seeds from real-world failure data. While this is a core feature of the original Surrealist for UAVs, feedback reaffirms its importance in reducing the high manual effort for seed creation. Also, a dedicated visualization tool to compare algorithm runs side-by-side and improved statistical metrics were suggested to enhance performance benchmarks.
\item \textit{Extension}: the framework's success in navigation has sparked interest in expanding its scope, e.g., handling more complex challenges—such as navigation without predefined waypoints—and extending it to new domains like locomotion testing.
\end{enumerate}

\subsection{Takeaways for Researchers and Robotic Practitioners} 
\begin{itemize}[leftmargin=*, topsep=0pt, itemsep=0pt, parsep=0pt]
    \item \textit{Integrate with a Non-Invasive Abstraction Layer:} The success of this project relied on ANYmal’s system test interface — a facade we developed to let our research prototypes (Surrealist and Aerialist) interact with ANYbotics' proprietary software stack in a black-box, non-invasive way. This was key for industrial adoption, as it avoided disrupting existing workflows.
    Additionally, Aerialis's generic test description (in YAML) and the framework's decoupled architecture enabled a smooth extension from UAVs to the quadrupeds. 

    \item \textit{Adopt a User-Driven, Iterative Development Process:} In addition to advancing core test generation algorithms, it is important for researchers to also consider the usability of testing tools and the practical relevance of their outputs to support broader industrial adoption. The pilot phase—with its close, iterative feedback loop involving \textit{ANYbotics} engineers —was instrumental in refining the framework. Direct user feedback led to meaningful improvements in visualization, performance metrics, and essential features such as test suite re-execution, ultimately enhancing the tool’s practical utility and enabling its successful real-world deployment. 
    
    \item \anybotics{\textit{Address the Sim-to-Real Gap Strategically:} The survey revealed the sim-to-real gap is a relative, not absolute, barrier. It was considered manageable for navigation, where the robot is modeled accurately, the environment is abstracted through sensor preprocessing, and movements are streamlined by robust locomotion. Here, the framework successfully predicted real-world algorithmic failures, building significant trust with the engineers. The primary concerns were perception-related (e.g., lack of sensor noise, oversimplified obstacles), which reinforces a practical workflow: use simulation for broad, cost-effective discovery of high-level algorithmic flaws, and reserve targeted physical testing for scenarios where perception performance may degrade.}

\end{itemize}

\subsection{Threats to Validity}
\label{sec:threats}

While our findings support the positive impact of the integrated framework, several threats to validity must be considered. The results are fundamentally dependent on the fidelity of the simulation environment and are subject to the well-known ``reality gap'': behaviors observed in simulation may not perfectly translate to the physical world. Our test generation process is rooted in a limited set of manually designed seed scenarios and is guided by a fitness function focused primarily on obstacle proximity. These choices, while effective, may not capture all relevant edge cases or dimensions of scenario difficulty. Similarly, the performance metrics used do not encompass all aspects of navigation quality, such as energy efficiency or motion smoothness. Finally, while the evaluation was conducted on an industrial-grade platform, the findings are specific to the ANYmal robot and the organizational context of a single company, and the qualitative insights are derived from a small group of eight engineers.

\section{Related Works}
\label{sec:relatedworks}

Testing of robotic systems—especially for autonomous navigation—remains difficult due to system complexity and unpredictable environments~\cite{afzal2020study}. Designing realistic test environments and oracles is notoriously challenging~\cite{afzal2020study,lindvall2017metamorphic}, limiting the adoption of simulation-based testing despite its benefits~\cite{afzal2021simulation}. The gap between simulated and real-world performance further complicates validation~\cite{afzal2021simulation}. Our work addresses these challenges by extending our prior work on Surrealist~\cite{surrealist}—a search-based approach for generating challenging test variants for UAVs—to quadrupedal robots and integrating it into an industrial workflow.

A significant body of research focuses on generating diverse environments to improve test coverage. Procedural content generation, as explored by Sotiropoulos et al.~\cite{sotiropoulos2017can}, uses randomization to create open-space worlds to find navigation bugs, with a focus on assessing the difficulty of these worlds~\cite{sotiropoulos2016virtual}. To provide more structure, Parra et al.~\cite{parra2023thousand} introduced FloorPlan DSL for defining indoor test environments and Variation DSL for structured variability, such as sampling obstacle sizes from a normal distribution. Similarly, tools like \textit{Local Planner Bench}~\cite{spahn2022local} support randomized environments to benchmark local obstacle avoidance algorithms. While these methods excel at creating a wide variety of test scenarios, they are generally undirected. In contrast, our approach uses a search-based algorithm to systematically and purposefully evolve scenarios to find failure-inducing configurations.

Another line of research focuses specifically on automated techniques to find failures. Fuzzing methods, such as PHYS-FUZZ by Woodlief et al.~\cite{elbaum2021fuzzing}, vary environmental parameters and robot poses to uncover crashes in mobile robots. This is conceptually similar to approaches in autonomous driving, where simulation is used to generate failure scenarios by altering the environment, traffic behavior, and sensor inputs~\cite{GambiMF19, stocco2020, VRstudy,DBLP:journals/ese/BirchlerKBGP23}.

Search-based techniques offer a more guided approach to failure discovery. For instance, Humeniuk et al.~\cite{humeniuk2024simulation} proposed MARTENS, a search-based method for testing the DL vision models used in autonomous manipulators. Our work also falls into this category, but while prior approaches often focus on specific input space exploration (fuzzing) or perception modules (MARTENS), our work targets the full-system behavior of an industrial quadrupedal robot. We use search to optimize for challenging dynamics in obstacle avoidance, offering a more comprehensive, system-level testing strategy. Our successful integration and evaluation on the ANYmal platform further highlight the real-world applicability of this approach for improving the robustness of complex legged robots.

\section{Conclusion and Future Work}
\label{sec:conlusion}
This paper successfully demonstrated how an academic, search-based testing framework can be adapted and integrated into a complex industrial robotics workflow. Our two-phased evaluation at ANYbotics confirmed the framework's value: it significantly streamlined the testing process, uncovered critical failures, and provided an objective means to benchmark performance improvements. This work serves as a case study for bridging the research-to-practice gap, delivering a mature, industrially-validated tool that enhances the robustness of the ANYmal robot and accelerates the overall development cycle.

Future work will focus on expanding the framework's capabilities based on the received feedback. Our primary research directions include enhancing scenario realism by incorporating more complex environmental features and sensor models, replicating real-world tests in simulation, and broadening the framework's applicability to new robotic domains.

\section*{Acknowledgments}
We thank the Horizon and SERI for supporting the \href{https://www.innoguard.eu/index.html}{InnoGuard} project (Marie Skłodowska-Curie DN, HORIZON-MSCA-2023-DN), the SNSF for the "SwarmOps" project (No. 200021\_219732), and the Hasler Foundation for  the projects  "Aerialist" (No. 200021\_219732) and 
"Safe2Fly" (No. 2025-02-27-311). 
\anybotics{We sincerely thank ANYbotics AG for their full support of this study, with special gratitude to Gabriel Hottiger, Rene Hölbling, and Yi Hao Ng for their supervision, and all survey participants.}
\bibliographystyle{IEEEtran}
\bibliography{references}
\end{document}